\documentclass[conference]{IEEEtran}
\IEEEoverridecommandlockouts
\usepackage{booktabs}
\usepackage{cite}
\usepackage{comment}
\usepackage{amsmath,amssymb,amsfonts}
\usepackage{algorithmic}
\usepackage{graphicx}
\usepackage{textcomp}
\usepackage{xcolor}
\usepackage{hyperref}
\usepackage{multirow}
\usepackage[linesnumbered,ruled,vlined]{algorithm2e}
\usepackage{tikz}
\def\BibTeX{{\rm B\kern-.05em{\sc i\kern-.025em b}\kern-.08em
    T\kern-.1667em\lower.7ex\hbox{E}\kern-.125emX}}
\usepackage{balance}

\hypersetup{
  colorlinks=true,
  linkcolor=blue,
  citecolor=blue,
  filecolor=magenta,
  urlcolor=blue
}

\makeatletter
\def\ps@IEEEtitlepagestyle{%
  \def\@oddfoot{\mycopyrightnotice}%
  \def\@evenfoot{}%
}
\def\mycopyrightnotice{%
  \gdef\mycopyrightnotice{}%
}
\makeatother

\usepackage{eso-pic}
\newcommand\AtPageUpperMyleft[1]{\AtPageUpperLeft{
 \put(\LenToUnit{1cm},\LenToUnit{-1cm}){ 
     \parbox{0.5\textwidth}{\raggedright\fontsize{9}{11}\selectfont #1}} 
 }}
\newcommand{\conf}[1]{
\AddToShipoutPictureBG*{
\AtPageUpperMyleft{#1}
}
}

\def\BibTeX{{\rm B\kern-.05em{\sc i\kern-.025em b}\kern-.08em
    T\kern-.1667em\lower.7ex\hbox{E}\kern-.125emX}}
\begin{document}

\title{Protecting Student Mental Health with a Context-Aware Machine Learning Framework for Stress Monitoring\\
}
\conf{This work has been submitted to the IEEE for possible publication. Copyright may be transferred without notice, after which this version may no longer be accessible.}

\author{\IEEEauthorblockN{
Md Sultanul Islam Ovi\IEEEauthorrefmark{1},
Jamal Hossain\IEEEauthorrefmark{3},
Md Raihan Alam Rahi\IEEEauthorrefmark{2}, and  
Fatema Akter\IEEEauthorrefmark{2}
}
\IEEEauthorblockA{\IEEEauthorrefmark{1}Dept. of Computer Science, George Mason University, Virginia, USA}
\IEEEauthorblockA{\IEEEauthorrefmark{3}Dept. of Computer Science and Engineering, University of Rajshahi, Bangladesh}
\IEEEauthorblockA{\IEEEauthorrefmark{2}Dept. of Computer Science and Engineering, Green University of Bangladesh, Bangladesh}
\IEEEauthorblockA{Email: movi@gmu.edu, 
jamal37.ru.cse@gmail.com,\\ raihanrahi170001@gmail.com, fatema.akter0313@gmail.com}
}

\maketitle

\begin{abstract}
Student mental health is an increasing concern in academic institutions, where stress can severely impact well-being and academic performance. Traditional assessment methods rely on subjective surveys and periodic evaluations, offering limited value for timely intervention. This paper introduces a context-aware machine learning framework for classifying student stress using two complementary survey-based datasets covering psychological, academic, environmental, and social factors. The framework follows a six-stage pipeline involving preprocessing, feature selection (SelectKBest, RFECV), dimensionality reduction (PCA), and training with six base classifiers: SVM, Random Forest, Gradient Boosting, XGBoost, AdaBoost, and Bagging. To enhance performance, we implement ensemble strategies including hard voting, soft voting, weighted voting, and stacking. Our best models achieve 93.09\% accuracy with weighted hard voting on the Student Stress Factors dataset and 99.53\% with stacking on the Stress and Well-being dataset, surpassing previous benchmarks. These results highlight the potential of context-integrated, data-driven systems for early stress detection and underscore their applicability in real-world academic settings to support student well-being.
\end{abstract}


\section{Introduction}

Mental health has become a significant concern in academic communities, with stress emerging as one of the most common psychological challenges affecting students. A growing body of evidence suggests that a substantial proportion of students experience moderate to high levels of stress each month, often driven by academic workload, social expectations, financial strain, and health-related issues \cite{Bryant2025}. Research further indicates that nearly one-third of students globally face serious mental health challenges exacerbated by sustained academic pressure \cite{Prez-Jorge2025}. If left unaddressed, prolonged stress may lead to adverse physical and psychological outcomes such as anxiety, depression, cognitive fatigue, and a general decline in quality of life \cite{Abd-Alrazaq2024}.

Traditional methods for stress detection rely primarily on subjective surveys, periodic interviews, or occasional biological assays such as cortisol measurements \cite{Gedam2021}. Instruments like the Perceived Stress Scale (PSS) are frequently used in academic research to assess stress perception. However, these tools are subject to recall bias, mood dependency, and participant fatigue \cite{Gedam2021}. Moreover, they provide only a retrospective snapshot of a student's mental state, limiting their usefulness for early intervention. While physiological assays offer more objective data, they are often time-intensive and impractical for continuous use in real-world settings \cite{Paniagua-Gmez2025}. As a result, students experiencing elevated stress levels may go unnoticed until symptoms become severe.

Recent technological advancements have introduced new opportunities for dynamic and scalable stress monitoring solutions. The integration of machine learning (ML) with Internet of Things (IoT) technologies, particularly wearable sensors and smartphones, has enabled continuous and unobtrusive monitoring of physiological and behavioral indicators of stress \cite{Paniagua-Gmez2025}. Devices such as smartwatches and fitness trackers can collect real-time biosignals including heart rate variability, electrodermal activity, and skin temperature \cite{Kallio2025}. These data streams are capable of reflecting autonomic nervous system activity that correlates with psychological stress \cite{Xiang2025}. In addition to these signals, smartphones and ambient sensors can provide contextual information such as physical activity, type of environment, time of day, and social setting \cite{Talaat2023}.

Many existing detection systems, however, do not incorporate contextual information, which limits their ability to distinguish between physiological responses driven by different underlying conditions \cite{Onim2024}. For instance, an elevated heart rate may result from either psychological stress or physical exertion. Integrating contextual data such as the user's activity and environment helps resolve such ambiguities and enhances detection accuracy \cite{Aqajari2023}. Studies have demonstrated that including contextual variables in stress prediction models yields significantly better performance than models relying on physiological signals alone \cite{Aqajari2023}.

In this study, we propose a context-aware machine learning framework for stress detection among students. Our approach combines two complementary survey-based datasets \cite{StudentStressFactors2024, StressWellBeingCollegeStudents2024} to formulate multiclass prediction tasks. The proposed framework is adaptive and personalized, designed to detect stress and enable early interventions. In practical applications, the system can provide alerts and feedback when elevated stress is identified, allowing students to take proactive measures to manage their mental health. Furthermore, aggregated insights from the system can inform institutional wellness strategies by identifying stress trends during critical academic periods.

\section{Literature Review}

Stress monitoring with machine learning has advanced significantly, particularly through the use of wearable sensors and context-aware systems. Wearable devices are now widely adopted due to their ability to continuously collect physiological signals without intruding on daily routines \cite{Pinge2024, Al-Atawi2023}. Devices such as smartwatches and wristbands equipped with sensors for heart rate variability, electrodermal activity, and skin temperature have been used to train machine learning models capable of detecting stress \cite{Xiang2025, Abd-Alrazaq2024}. Public datasets like WESAD support this work, although limited recording durations and small cohorts often hinder generalization \cite{Vos2023generalizable}. To overcome this, recent studies like Vos et al. (2023) combined data from multiple sources and used ensemble models, achieving up to 85 percent accuracy on unseen data \cite{Vos2023ensemble}. Deep learning has also been applied effectively. Xiang et al. (2025) used a convolutional neural network to detect stress in healthcare workers with an F1-score around 0.91 \cite{Xiang2025}. Despite this progress, challenges remain, including missing data in free-living conditions, individual variability in physiological baselines, and privacy concerns \cite{Onim2024, Su2024}.

Survey-based approaches provide a complementary angle by capturing subjective experiences. Instruments such as the Perceived Stress Scale and Depression Anxiety Stress Scales remain widely used \cite{Gedam2021, Firoz2023}. Anand et al. (2023) demonstrated that survey responses about academic workload and lifestyle could be used to classify stress levels with over 93 percent accuracy using ensemble classifiers \cite{Anand2023}. Class imbalance was addressed through oversampling, and the model's performance remained stable across folds. Survey data are valuable for identifying individual stressors and validating physiological models, although issues such as recall bias and inconsistent participation persist. Ecological momentary assessments have been explored to gather real-time subjective stress labels, offering temporally rich training data \cite{Aqajari2023, Onim2024, singh2024machine}.

Context-aware stress detection has emerged as a promising improvement over models that rely solely on physiological inputs \cite{Singh2024distress}. Aqajari et al. (2024) combined wearable photoplethysmography with smartphone-based context data and achieved an F1-score of approximately 0.70, compared to 0.56 with physiological data alone \cite{Aqajari2023}. Contextual inputs such as activity, environment, and time-of-day provide disambiguation for ambiguous signals like elevated heart rate. The StudentLife study exemplified the use of smartphones for behavior monitoring in student populations. These approaches improve prediction but raise privacy challenges and require careful integration of heterogeneous data streams \cite{Onim2024, Su2024}.

AI chatbots represent an intervention-focused line of research that complements detection. Chatbots like Woebot and Wysa use natural language processing to guide users through coping exercises and have shown positive mental health outcomes in student populations \cite{Umashankar2024, Bhole2018}. Liu et al. (2022) found that students using chatbots reported reductions in depressive symptoms \cite{Joshi2022}. Chatbots can be integrated into monitoring systems for real-time support and may also serve as a data source for emotion detection. However, concerns remain regarding long-term engagement, safety, and limitations in emotional understanding \cite{Umashankar2024}. These tools are best used alongside human support services and as part of a broader mental health strategy.

Ensemble and deep learning techniques have gained traction in stress detection. Studies have shown that combining multiple models improves generalizability and stability. Anand et al. (2023) and Vos et al. (2023) both demonstrated that ensemble classifiers outperform single models in both survey and wearable-based contexts \cite{Anand2023, Vos2023ensemble, Jayawickrama2022}. Deep models such as convolutional encoders and multimodal LSTMs further boost performance by capturing complex temporal patterns \cite{Xiang2025, Onim2024}. Onim et al. (2024) used a CNN-based encoder to extract features from wearable data, achieving comparable or better results than hand-crafted features \cite{Onim2024}. Though deep models require large datasets and pose interpretability challenges, they represent a key direction for future research \cite{Vos2023generalizable, Paniagua-Gmez2025}.

Despite these advancements, key gaps remain. Many models still lack context integration, limiting their real-world applicability \cite{Onim2024}. Ensemble learning is underused relative to its potential, and generalizability across diverse populations remains limited \cite{Vos2025stress}. Data scarcity and class imbalance are persistent issues \cite{Anand2023, Gedam2021}. Furthermore, most systems focus solely on detection without linking to intervention, and few address privacy at the design level \cite{Su2024}.

Our proposed framework responds to these gaps by combining comprehensive survey-based features across psychological, physiological, environmental, academic, and social dimensions through ensemble machine learning approaches. The framework integrates two complementary datasets and is designed to be adaptive and student-centric. By building on recent findings in survey-based stress detection, our system aims to deliver improved accuracy and support proactive mental health strategies in educational environments.

 \begin{figure*}[htbp]
   \begin{center}
        \includegraphics[width=\linewidth]{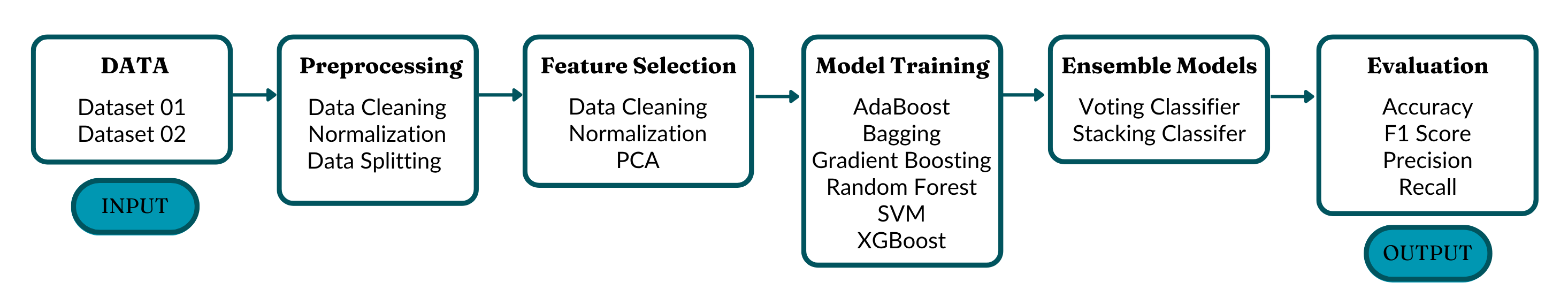}
    \end{center}
  \caption{Context-Aware Machine Learning Framework for Student Stress Classification}
  \label{fig:framewo}
\end{figure*}

\section{Framework Overview}

The proposed system classifies student stress levels through multi-modal survey data analysis, as shown in Figure~\ref{fig:framewo}. The six-stage pipeline includes data preprocessing (missing value handling and normalization), train-test splitting, feature selection (SelectKBest, RFECV), and dimensionality reduction (PCA with 90\%, 95\%, and 99\% variance retention). It then trains base classifiers (SVM, Random Forest, Gradient Boosting, XGBoost, AdaBoost, Bagging) and integrates ensemble learning methods (hard voting, soft voting, weighted voting, and stacking). Each stage is evaluated using accuracy, F1-score, precision, and recall. The ensemble models combine the strengths of individual classifiers to ensure robust performance for real-world stress classification. The pipeline’s implementation is detailed in Algorithm~\ref{tab:algo}.

\subsection{Data Collection and Dataset Description}

The evaluation uses two datasets designed for student stress analysis, each capturing distinct aspects of stress and providing complementary perspectives on student mental health. Table \ref{tab:datasets} summarizes the key characteristics of each dataset.

\begin{table}[htbp]
\centering
\caption{Summary of the Datasets Used for Stress Classification}
\label{tab:datasets}
\begin{tabular}{|l|c|c|}
\hline
\textbf{Dataset Name} & \textbf{Total Samples} & \textbf{Number of Features} \\ \hline
Student Stress Factors\cite{StudentStressFactors2024} & 1,100 & 21 \\ \hline
Stress \& Well-being Data\cite{StressWellBeingCollegeStudents2024} & 843 & 26 \\ \hline
\end{tabular}
\end{table}

The \textit{Student Stress Factors: A Comprehensive Analysis}\cite{StudentStressFactors2024} dataset includes 1,100 records across 21 features, covering psychological, physiological, environmental, academic, and social stress factors. The \textit{Stress and Well-being Data of College Students}\cite{StressWellBeingCollegeStudents2024} dataset contains 843 records with 26 features, focusing on stress indicators such as sleep issues, anxiety, and academic workload. Both datasets were subjected to quality assessments, including missing value analysis, duplicate detection, and statistical profiling, ensuring robust evaluation of the framework across different stress assessment methodologies.

\begin{algorithm}
\label{tab:algo}
\caption{ML Pipeline: Preprocessing, Feature Selection, Training, Evaluation}
\begin{algorithmic}[1]
\STATE \textbf{Input:} Dataset $D = (X, y)$
\STATE \textbf{Output:} Accuracy, F1 Score, Precision, Recall for all models

\STATE \textbf{Step 1: Preprocessing}
\STATE Handle missing values in $X$
\STATE Normalize features: $X \leftarrow$ scaler.fit\_transform($X$)

\STATE \textbf{Step 2: Train-Test Split}
\STATE Split into training and testing sets: $X_{\text{train}}, X_{\text{test}}, y_{\text{train}}, y_{\text{test}}$

\STATE \textbf{Step 3: Feature Selection}
\STATE Apply SelectKBest on $X_{\text{train}}$: $X_{\text{train\_kbest}} \leftarrow$ kbest.fit\_transform($X_{\text{train}}, y_{\text{train}}$)
\STATE Transform test set: $X_{\text{test\_kbest}} \leftarrow$ kbest.transform($X_{\text{test}}$)

\STATE Apply RFECV on $X_{\text{train\_kbest}}$: $X_{\text{train\_rfecv}} \leftarrow$ rfecv.fit\_transform($X_{\text{train\_kbest}}, y_{\text{train}}$)
\STATE Transform test set: $X_{\text{test\_rfecv}} \leftarrow$ rfecv.transform($X_{\text{test\_kbest}}$)

\STATE \textbf{Step 4: Dimensionality Reduction (PCA)}
\STATE Apply PCA: $X_{\text{train\_pca}} \leftarrow$ pca.fit\_transform($X_{\text{train\_rfecv}}$)
\STATE Transform test set: $X_{\text{test\_pca}} \leftarrow$ pca.transform($X_{\text{test\_rfecv}}$)

\STATE \textbf{Step 5: Base ML Model Training and Evaluation}
\FOR{each base model $m \in \{$SVM, Random Forest, Gradient Boosting, XGBoost, AdaBoost, Bagging$\}$}
    \STATE Train $m$ on appropriate feature set (e.g., $X_{\text{train\_pca}}$ or $X_{\text{train\_rfecv}}$)
    \STATE Predict: $\hat{y}_m \leftarrow m.predict(X_{\text{test}})$
    \STATE Compute Accuracy, F1 Score, Precision, Recall
\ENDFOR

\STATE \textbf{Step 6: Ensemble Model Construction and Evaluation}
\STATE Build Voting and Stacking classifiers using trained base models
\FOR{each ensemble model $e \in \{$Voting (hard, soft, weighted), Stacking$\}$}
    \STATE Train $e$ on $X_{\text{train\_mixed}}$, $y_{\text{train}}$
    \STATE Predict: $\hat{y}_e \leftarrow e.predict(X_{\text{test\_mixed}})$
    \STATE Compute Accuracy, F1 Score, Precision, Recall
\ENDFOR

\RETURN All evaluation metrics for base and ensemble models
\end{algorithmic}
\end{algorithm}

\subsection{Preprocessing and Feature Engineering}

The preprocessing pipeline includes data quality assessments such as missing value imputation, duplicate removal, and normalization using MinMaxScaler, which scales features to a [0,1] range. Feature selection strategies optimize model performance: SelectKBest identifies the most informative features via F-classification testing, with the optimal count determined through cross-validation with SVM classifiers. Recursive Feature Elimination with Cross-Validation (RFECV) removes the least important features using a linear SVM as the base estimator and 5-fold stratified cross-validation. Principal Component Analysis (PCA) is applied to reduce dimensionality while retaining 90\%, 95\%, and 99\% of the variance. These steps result in five distinct preprocessing configurations used to evaluate all base classifiers.

\begin{table*}[htbp]
\caption{Evaluation Metrics (\%) Across Two Datasets}
\centering
\begin{tabular}{|c|c|cccc|cccc|}
\hline
\multirow{2}{*}{\textbf{ML Model}} 
& \multirow{2}{*}{\textbf{Configuration}} 
& \multicolumn{4}{c|}{\textbf{Dataset 1}} 
& \multicolumn{4}{c|}{\textbf{Dataset 2}} \\
\cline{3-10}
& & \textbf{Accuracy} & \textbf{F1 Score} & \textbf{Recall} & \textbf{Precision} 
 & \textbf{Accuracy} & \textbf{F1 Score} & \textbf{Recall} & \textbf{Precision}  \\
\hline
AdaBoost                          & SelectKBest         & 92.000 & 92.005 & 92.083 & 92.121 & 96.683 & 85.866 & 79.372 & 94.818 \\
\hline
Bagging                           & SelectKBest         & 89.455 & 89.493 & 89.508 & 89.716 & 94.313 & 68.831 & 59.091 & 98.039 \\
\hline
Gradient Boosting                 & SelectKBest         & 91.273 & 91.279 & 91.404 & 91.498 & 95.735 & 79.044 & 69.318 & 98.508 \\
\hline
Random Forest                     & SelectKBest         & \textbf{92.364} & 92.365 & 92.356 & 92.378 & 94.787 & 72.332 & 63.258 & 98.194 \\
\hline
Support Vector Machine            & PCA                 & 90.546 & 90.557 & 90.551 & 90.567 & \textbf{99.052} & 96.494 & 93.939 & 99.656 \\
\hline
XGBoost                           & SelectKBest         & 91.636 & 91.636 & 91.776 & 92.014 & 96.209 & 82.058 & 73.485 & 98.667 \\
\hline

\end{tabular}
\label{tab:combined_metrics}
\end{table*}

\subsection{Machine Learning Algorithms}

The framework employs six machine learning algorithms, each suited for different aspects of stress classification. Support Vector Machine (SVM) handles high-dimensional data with strong generalization. Random Forest provides ensemble learning benefits and feature importance ranking. Gradient Boosting optimizes weak learners sequentially, improving performance on misclassified instances. AdaBoost adjusts weights to enhance performance on difficult instances. XGBoost extends gradient boosting with regularization and computational efficiency. Bagging Classifier reduces variance through bootstrap aggregating, ensuring stable predictions across diverse data distributions.

\subsection{Customized Ensemble Design}

The ensemble learning framework implements five aggregation strategies to leverage complementary strengths of base classifiers. Hard Voting Classifier employs majority vote decision-making, while Soft Voting Classifier utilizes probability-weighted predictions from all base models. Weighted Hard Voting assigns higher weights to superior-performing models in the majority voting process. Weighted Soft Voting extends probability-based voting with performance-weighted averaging based on individual model strengths. Stacking Classifier employs a meta-learner to optimally combine base model predictions. This method trains a secondary model using base classifier predictions as input features, learning the optimal combination strategy through cross-validated training to prevent overfitting while maximizing ensemble performance.

\subsection{Hyperparameter Tuning}

Hyperparameter tuning is conducted using GridSearchCV with 10-fold cross-validation. The search space includes key parameters for each model, such as SVM regularization strength, kernel type, and gamma, along with algorithm-specific settings for tree-based and boosting methods. Base model parameters are tuned individually before ensemble combination, with parallel processing utilized to efficiently explore the hyperparameter space while maintaining computational feasibility.

\subsection{Training and Validation Protocol}

The experimental protocol implements a 75\%-25\% stratified train-test split to ensure balanced class representation. Cross-validation employs 10-fold Stratified K-Fold for feature selection and hyperparameter optimization, and 5-fold for model evaluation. All random processes use seed value 42 for reproducibility. The framework systematically evaluates each base classifier across all preprocessing configurations: original data, normalized features, SelectKBest-selected features, RFECV-selected features, and PCA-transformed components. This enables identification of optimal preprocessing-model combinations.

\subsection{Evaluation Metrics}

Model performance is assessed using four evaluation metrics with macro-averaging to ensure equitable treatment across all stress categories. Accuracy measures overall classification correctness, serving as the primary performance indicator. Precision quantifies positive prediction accuracy, while Recall evaluates true positive identification rate. The F1-score is the harmonic mean of precision and recall, making it useful for evaluating performance when dealing with imbalanced or multiclass classification tasks.

\section{Results and Discussion}
\subsection{Performance of Individual Classifiers}

Individual classifiers achieved strong baseline performance across both datasets. Table~\ref{tab:combined_metrics} presents the comprehensive evaluation metrics for all base classifiers. On Dataset 1 (Student Stress Factors), Random Forest with SelectKBest preprocessing achieved the highest accuracy of 92.364\%, demonstrating the effectiveness of feature selection for tree-based algorithms. Support Vector Machine with PCA preprocessing reached 90.546\% accuracy, indicating successful dimensionality reduction while maintaining classification performance. Dataset 2 (Stress and Well-being Data) showed remarkable performance, with Support Vector Machine achieving 99.052\% accuracy using PCA preprocessing. This suggests that the multiclass classification task benefits significantly from dimensionality reduction techniques, enabling SVM to effectively separate stress categories in the transformed feature space. Gradient Boosting, AdaBoost, and XGBoost demonstrated consistent performance, with AdaBoost achieving 92.000\% accuracy on Dataset 1 and 96.683\% on Dataset 2. The ensemble nature of these algorithms proved effective for stress classification tasks, capturing complex non-linear relationships between stress indicators.

\begin{table}[htbp]
\caption{Accuracy (\%) of Machine Learning and Ensemble Models on Dataset~1 and Dataset~2}
\centering
\begin{tabular}{|l|c|c|}
\hline
\textbf{ML \& Ensemble Model} & \textbf{Dataset 01} & \textbf{Dataset 02} \\
\hline
AdaBoost & 92.000 & 96.683 \\
\hline
Bagging & 89.455 & 94.313 \\
\hline
Gradient Boosting & 91.273 & 95.735 \\
\hline
Random Forest & 92.364 & 94.787 \\
\hline
Support Vector Machine & 90.546 & 99.052 \\
\hline
XGBoost & 91.636 & 96.209 \\
\hline
Voting Classifier (hard) & \textbf{93.091} & 97.156 \\
\hline
Voting Classifier (soft) & 92.364 & 97.630 \\
\hline
Voting Classifier (weighted\_hard) & \textbf{93.091} & 99.052 \\
\hline
Voting Classifier (weighted\_soft) & 92.364 & 99.052 \\
\hline
Stacking Classifier & 91.273 & \textbf{99.530} \\
\hline
\end{tabular}
\label{tab:accuracies_only}
\end{table}

\subsection{Effectiveness of the Ensemble Model}

Ensemble methods outperformed individual classifiers, as demonstrated in Table~\ref{tab:accuracies_only}. On Dataset 1, Hard Voting Classifier and Weighted Hard Voting Classifier both achieved 93.091\% accuracy, representing a 0.727\% improvement over the best individual classifier (Random Forest at 92.364\%). For Dataset 2, the Stacking Classifier achieved the highest accuracy of 99.530\%, followed by Weighted Soft Voting, Weighted Hard Voting, and SVM, each with 99.052\%. These results indicate that the meta-learning approach in stacking was effective in identifying optimal combination strategies for the multiclass classification task. Weighted voting strategies also showed strong performance, suggesting that performance-based weighting can offer advantages over simple majority voting.

\subsection{Comparative Study with Existing Works}

To evaluate the effectiveness of our models, we compared them with prior studies that used the same datasets. For Dataset~1 (Student Stress Factors), the Springer chapter by Doma et al.~\cite{doma2025classification} achieved 92.41\% accuracy using a hybrid ensemble. De Filippis and Foysal~\cite{de2024comprehensive} and Amalia et al.~\cite{amalia2025optimization} both reported 88.0\% using Random Forest, with and without feature selection. Our Weighted Voting Classifier reached 93.09\% accuracy, exceeding all prior works. Even our best single model (Random Forest with SelectKBest) slightly outperformed Doma et al.'s ensemble. 

\begin{figure}[htbp]
   \centering
   \includegraphics[width=.8\linewidth]{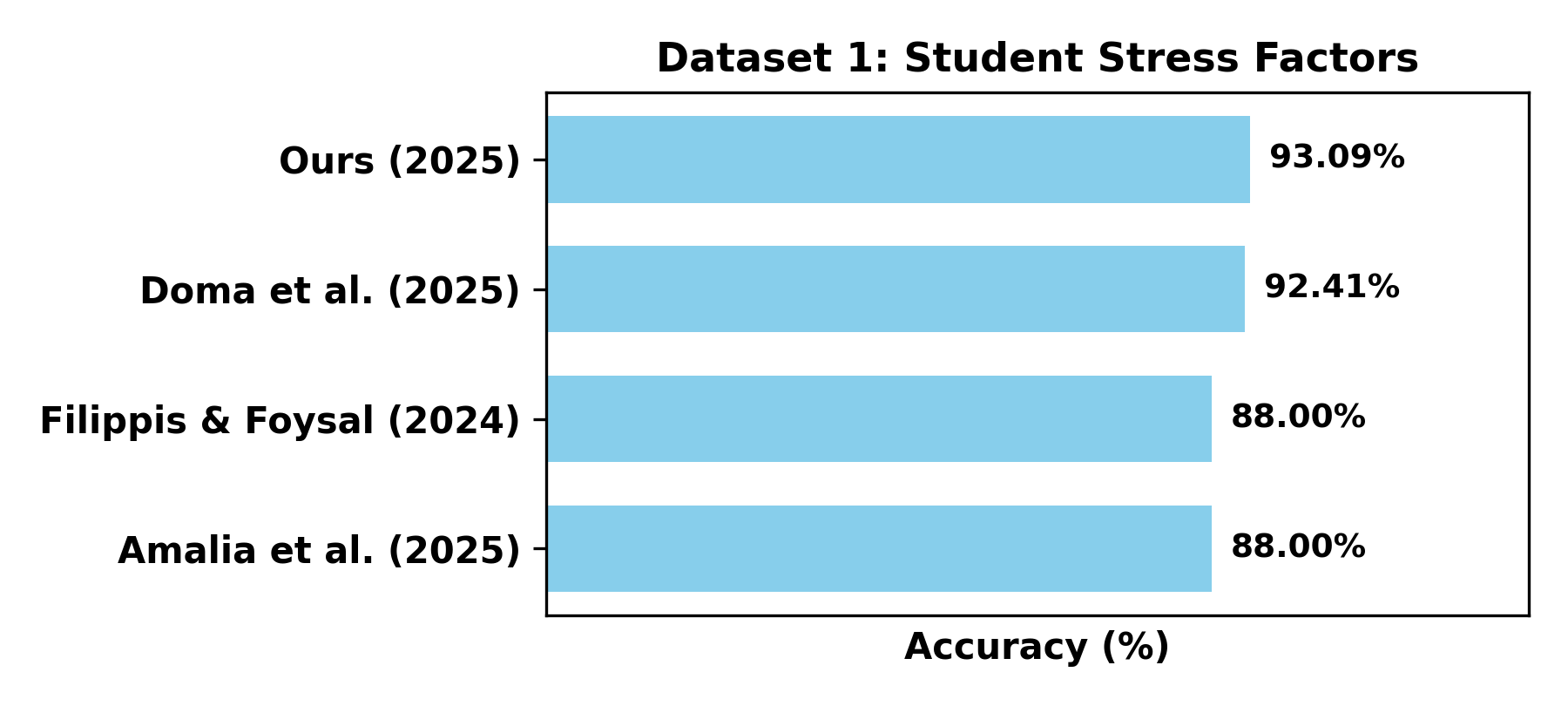}
   \caption{Accuracy comparison with existing models on Dataset~1.}
   \label{fig:dataset1_comparison}
\end{figure}

\begin{figure}[htbp]
   \centering
   \includegraphics[width=.8\linewidth]{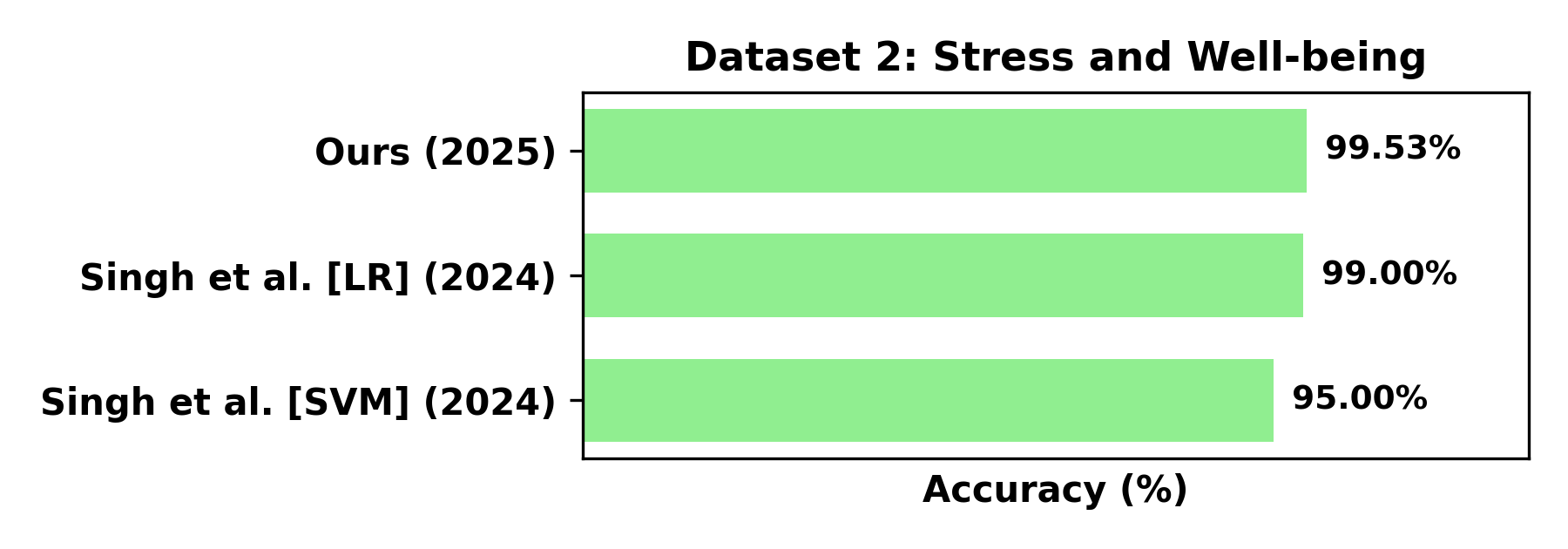}
   \caption{Accuracy comparison with existing models on Dataset~2.}
   \label{fig:dataset2_comparison}
\end{figure}

For Dataset~2 (Stress and Well-being), Singh et al.~\cite{singh2024machine} reported 95\% accuracy using SVM, and in a related study~\cite{singh2024optimizing}, Logistic Regression achieved 99\% on distress classification. Our SVM with PCA reached 99.05\%, and the Stacking Classifier achieved 99.53\%, outperforming all previous results. Figures~\ref{fig:dataset1_comparison} and~\ref{fig:dataset2_comparison} present separate visual comparisons for each dataset, clearly illustrating the superior performance of our proposed models.

\section{Ethical Considerations and Privacy}

The framework development prioritizes ethical considerations and privacy protection throughout the research process. All datasets utilized are publicly available and anonymized, ensuring no personally identifiable information is processed. The research adheres to established ethical guidelines for machine learning research in healthcare applications. Potential risks include algorithmic bias toward specific demographic groups and over-reliance on automated detection without human oversight. The framework mitigates these risks through comprehensive evaluation across diverse datasets and multi-metric assessment to prevent bias toward majority classes. Future deployment considerations include obtaining appropriate student consent, implementing robust data anonymization procedures, ensuring algorithm transparency for stakeholders, and establishing clear protocols for responding to high-stress classifications. The framework is designed to support rather than replace human mental health professionals, emphasizing the importance of human judgment in student well-being decisions.

\section{Conclusion}

This paper introduced a context-aware machine learning framework for student stress classification, built on two complementary survey datasets capturing psychological, academic, and lifestyle stressors. The proposed system integrates preprocessing, feature selection (SelectKBest, RFECV), dimensionality reduction (PCA), and six base classifiers, combined with five ensemble strategies including weighted voting and stacking. Experimental results show that ensemble models consistently outperform individual classifiers, with the weighted hard voting classifier achieving 93.09\% accuracy on Dataset~1 and the stacking classifier reaching 99.53\% on Dataset~2. These results exceed prior benchmarks and highlight the importance of combining diverse models with proper preprocessing. The framework’s strong performance and adaptability suggest its potential for real-world deployment in academic environments to support early mental health interventions. Future work will focus on integrating real-time physiological and behavioral data, developing personalized adaptation mechanisms, and exploring privacy-preserving deployment strategies such as federated learning.

\bibliography{references}

\end{document}